\documentclass[aps,prl,twocolumn,groupedaddress]{revtex4}
\usepackage{amsfonts}
\usepackage{mathrsfs}
\usepackage{graphicx}
\usepackage{subfigure}
\usepackage{amsmath}
\usepackage{amssymb}
\usepackage{amsbsy}
\usepackage{bm}

\def\E{\mathbb{E}}

\begin{document}

\title{Learning credit assignment}
\author{Chan Li}
\author{Haiping Huang}
\affiliation{PMI Lab, School of Physics,
Sun Yat-sen University, Guangzhou 510275, People's Republic of China}
\date{\today}

\begin{abstract}
Deep learning has achieved impressive prediction accuracies in a variety of scientific and industrial domains. However, the nested non-linear feature of deep learning 
makes the learning highly non-transparent, i.e., it is still unknown how the learning coordinates a huge number of parameters to achieve a decision making.
To explain this hierarchical credit assignment, we propose a mean-field learning model by assuming that an ensemble of sub-networks, rather than a single network, are trained for 
a classification task. Surprisingly, our model reveals that apart from some deterministic synaptic weights connecting two neurons at neighboring layers, there exist a large number of 
connections that can be absent, and other connections can allow for a broad distribution of their weight values. Therefore, synaptic connections can be classified into three categories: very important ones, unimportant ones, and 
those of variability that may partially encode nuisance factors. Therefore, our model learns the credit assignment leading to the decision, and predicts an ensemble of sub-networks that can accomplish the 
same task, thereby providing insights toward understanding the macroscopic behavior of deep learning through the lens of distinct roles of synaptic weights.
\end{abstract}

 \maketitle

\textit{Introduction.}---
As deep neural networks become an increasingly important tool in diverse domains of scientific and engineering applications~\cite{Hinton-2012,Dsp-2014,He-2015,Silver-2017,DLNS-2019},
the black box properties of the tool turn out to be a challenging obstacle puzzling researchers in the field~\cite{Neuron-2019}. In other words, the decision making behavior of a network 
output can not be easily understood in terms of interactions among building components of the network. This shares the same spirit as another long-standing puzzle in theory of the brain---how the emergent behavior
of a neuronal population hierarchy can be traced back to its elements~\cite{Neuron-2019,DLNS-2019}. This challenging issue is the well-known credit assignment problem (CAP), determining how much credit a given component (either 
neuron or connection) should take for a particular behavior output~\cite{DLNS-2019}. To solve this problem, one needs to bridge the gap between microscopic interactions of components and macroscopic behavior. 

Excitingly, recent works showed that there exist sub-networks of random weights that are able to produce better-than-chance accuracies~\cite{LTH-2018,Zhou-2019,Hidden-2019}. This property seems to be universal across different
architectures, datasets and computational tasks~\cite{Ticket-2019}. Even one can start with no connections and add complexity as needed by assuming a single shared weight parameter~\cite{Weight-2019}.
Moreover, it was recently revealed that
the innate template of face-selective neurons can spontaneously emerge from sufficient statistical variations present in the random initial wirings of neural 
circuits, while the template may be fine-tuned during early visual experiences~\cite{FSN-2019}. Therefore, from both artificial and biological neural networks perspectives,
exploring how and why these random wirings exist will definitely provide us a powerful lens through which we can better understand and even further improve the computational capacities of deep neural networks.

Here, we propose a statistical model of learning credit assignment from training data of a computational task. 
According to the model, we search for an optimal random network ensemble as an inductive bias about the hypothesis space~\cite{Neuron-2019}. The hypothesis space is composed of all candidate
networks with different assignments of
weight values that accomplish the computation task. Consistent with previous studies~\cite{LTH-2018,Zhou-2019,Hidden-2019}, the optimal ensemble
contains sub-networks of the original 
full network, which further allows for capturing uncertainty in the hypothesis space. The model can be solved by mean-field methods, thereby providing a physics interpretation of 
how credit assignment occurs in a hierarchical deep neural system.

\begin{figure}
\centering
     \includegraphics[bb=16 94 620 301,width=0.5\textwidth]{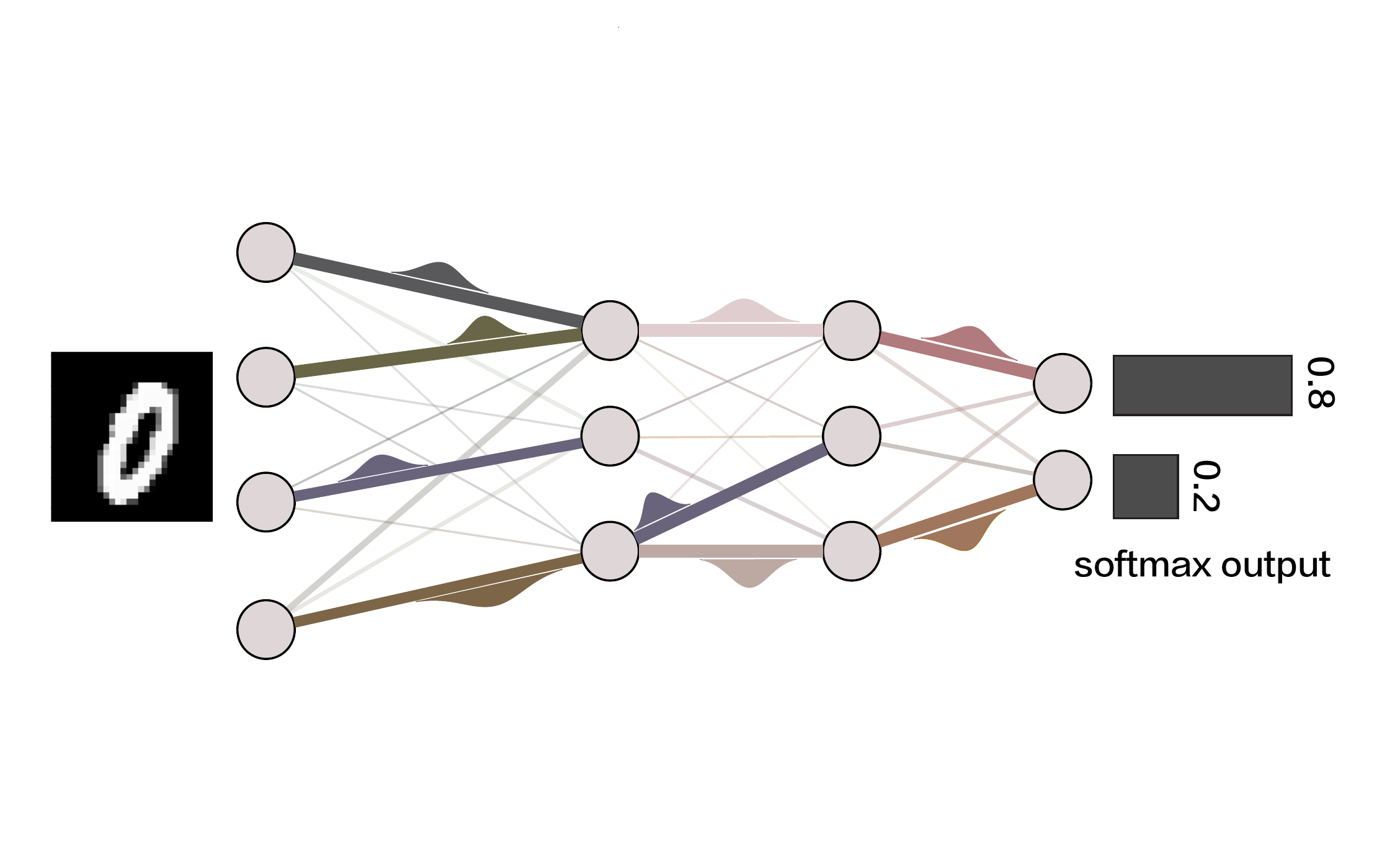}
  \caption{
  The schematic illustration of the model learning credit assignment.
 A deep neural network of four layers including two hidden layers is used to recognize a handwritten digit, say zero, 
 with the softmax output indicating the probability of the categorization. Each connection is specified by a spike and slab distribution, where 
 the spike indicates the probability of the absence of this connection, and the slab is modeled by a Gaussian distribution of weight values as pictorially shown only
 on strong connections with
 different means and variances. Other weak connections indicate nearly unit spike probabilities, although they also carry a slab
 distribution (not shown in the illustration for simplicity).
  }\label{model}
\end{figure}

\textit{Model.}---
To learn credit assignment, we search for an optimal random neural network ensemble to accomplish a classification task of handwritten digits~\cite{mnist}. More precisely,
we design a deep neural network of $L$ layers, including $L-2$ hidden layers. The depth of the network $L$ can be made arbitrarily large. The size (width) of the $l$-th layer is denoted 
by $N_l$, therefore $N_1$ is determined by the total number of pixels in an input image, and $N_L$ is the number of classes (e.g., $10$ for the 
handwritten digit dataset). The weight value of the connection from neuron $i$ at the upstream layer $l$ to neuron $k$ at the downstream layer $l+1$ is defined 
by $w_{ik}^l$, and the activation of the neuron $k$ at the $(l+1)$-th layer $h_k^{l+1}$ is a non-linear function of the pre-activation $z_k^{l+1}=\frac{1}{\sqrt{N_l}}\sum_iw_{ik}^lh_i^{l}$, where the weight scaling factor $1/\sqrt{N_l}$ ensures that
the weighted sum is independent of the upstream layer width.
We use the rectified linear unit (ReLU) function~\cite{Nair-2010} as the transfer function from pre-activation $z$ to activation $h$ defined as $h={\rm max}(0,z)$. The output transfer function specifies 
a probability over all classes of the input image by using a softmax function $h_k=\frac{e^{z_k}}{\sum_ie^{z_i}}$ where $z_i$ is the pre-activation of the $i$-th neuron at the output layer.
For the categorization task, we define $\hat{h}_i$ as the target label (one-hot representation), and
use the cross entropy $\mathcal{C}=-\sum_i\hat{h}_i\ln h_i$ as the objective function to be minimized.

Training the neural network corresponds to adjusting all connection weights to minimize the cross entropy until the network is able to classify unseen handwritten digits with
a satisfied accuracy (so-called generalization ability). Therefore, after the network is trained with a training data size of $T$, the network's generalization ability is verified 
with a test data size of $V$. Remarkably, the state-of-the-art test accuracy is able to surpass the human performance in some complex tasks~\cite{He-2015}. However, the decision making behavior
of the output neurons in a deep network is still challenging to understand in terms of computational principles of single building components (either neurons or weights).
Recent empirical machine learning works showed that a subnetwork of random weights can produce a better-than-chance accuracy~\cite{LTH-2018,Hidden-2019,Zhou-2019}.
This clearly suggests that there may exist a random ensemble of neural networks that fulfill the computational task given the width and depth of the deep network. This ensemble may occupy a tiny portion of the entire model space.
Therefore, a naive random initialization of the neural network can only yield a chance-level accuracy. To incorporate all these challenging issues into a theoretical model, we propose
to model the weight by a spike and slab (SaS) distribution as follows:

\begin{equation}\label{sas}
 P(w_{ik}^l)=\pi_{ik}^l\delta(w_{ik}^l)+(1-\pi_{ik})\mathcal{N}(w_{ik}^l|m_{ik}^l,\Xi_{ik}^l),
\end{equation}
where the discrete probability mass at zero defines the spike, and the slab is characterized by a Gaussian distribution with mean $m_{ik}^l$ and variance $\Xi_{ik}^l$ over a continuous domain (see Fig.~\ref{model} for an illustration).

The SaS distribution has been widely used in statistics literatures~\cite{sas-1988,Rao-2005}. Here, the spike and slab have respectively their own physics interpretations in learning
credit assignment of neural networks. The spike is intimately related to the concept of network compression~\cite{Han-2015,Max-2017,Ticket-2019}, where not all resources of connections are used in 
a task. This parameter allows to identify very important weights, and further evaluate remaining capacities for learning new tasks~\cite{CLL-2019}. A recent physics study has already showed that a deep 
neural network can be robust against connection removals~\cite{Rap-2018}. The slab continuous support
characterizes the ensemble of neural networks with random weights producing better-than-chance accuracies. Among these weights, some are very important, indicated by a vanishing spike probability mass, and could thus
explain the decision making of the output neurons, while the variance of the corresponding Gaussian distribution captures the uncertainty of the decision making solutions~\cite{Krieg-2016}.
Therefore, the inductive bias of connections and their associated weights can be learned through the SaS model of credit assignment. Note that the Gaussian slab is not used here as an additional
regularization complexity term in the objective function~\cite{Nowlan-1992}. Instead, the continuous slab is combined coherently with the spike probability to model the uncertainty of weights facing noisy sensory inputs (Fig.~\ref{model}).

Next, we derive a mean-field method to learn the SaS parameters $\bm{\theta}_{ik}^l\equiv(\pi_{ik}^l,m_{ik}^l,\Xi_{ik}^l)$ for all layers. The first and second moments of the weight $w_{ik}^l$ are given by
$\mu_{ik}^l\equiv\E[w_{ik}^l]=m_{ik}^l(1-\pi_{ik}^l)$ and $\varrho_{ik}^l\equiv\E[(w_{ik}^l)^2]=(1-\pi_{ik}^l)[\Xi_{ik}^l+(m_{ik}^l)^2]$, respectively. Given a large width of the layer, the central-limit-theorem implies that
the pre-activation follows approximately a Gaussian distribution 
$\mathcal{N}(z_i^l|G_i^l,(\Delta_i^l)^2)$, where the mean and variance are given respectively by
\begin{subequations}\label{mandv}
\begin{align}
 G_i^l&=\frac{1}{\sqrt{N_{l-1}}}\sum_{k}\mu_{ki}^{l-1}h_k^{l-1},\\
 (\Delta_i^l)^2&=\frac{1}{N_{l-1}}\sum_{k}(\varrho_{ki}^{l-1}-(\mu_{ki}^{l-1})^2)(h_k^{l-1})^2.
 \end{align}
\end{subequations}
Then, the feedforward transformation of the input signal can be re-parametrized by~\cite{LRT-2017,Huang-2019data}
\begin{subequations}\label{rep}
\begin{align}
 z_i^l&=G_i^l+\epsilon_i^l\Delta_i^l,\\
 h_i^l&={\rm ReLU}(z_i^l),
 \end{align}
\end{subequations}
for $l<L$. The last layer uses the softmax function. $\epsilon_i^l$ is a layer- and component-dependent standard Gaussian random variable with zero mean and unit variance.
$\boldsymbol{\epsilon}^l$ is quenched for every single training mini-epoch and the same value is used in both forward and backward computations.
This reparametrization retains the statistical structure in Eq.~(\ref{mandv}). The objective function relies on $(\bm{\mu},\bm{\varrho})$ that can be preserved by
a transformation of $\bm{\theta}$. Hence, we use a regularization strength to control the $\ell_2$-norm of $\bm{m}$ and $\bm{\Xi}$. In addition,
the transformation does not change the fraction of $\pi=1$ or $0$~\cite{SM}, maintaining the qualitative behavior of the model.

Learning of the hyper-parameter $\bm{\theta}_{ik}^l$ can be achieved by a gradient descent of the objective function, i.e.,
\begin{equation}\label{leq}
 \Delta\bm{\theta}_{ki}^l=-\eta\mathcal{K}_i^{l+1}\frac{\partial z_i^{l+1}}{\partial\bm{\theta}_{ki}^l}
\end{equation}
where $\eta$ denotes the learning rate, and $\mathcal{K}_{i}^{l+1}\equiv\frac{\partial\mathcal{C}}{\partial z_i^{l+1}}$.
The entire dataset is divided into minibatches over which the gradients are evaluated.
Note that unlike the standard back-propagation (BP)~\cite{Back-1986}, we adapt the SaS distribution rather than a
particular weight, in accord with our motivation of learning statistical features of the hypothesis space for a particular computation task (e.g., image classification here).

On the top layer, $\mathcal{K}_i^L$ can be directly estimated as $\mathcal{K}_i^L=-\hat{h}_i^{L}(1-h_i^L)$ by definition. For lower layers, $\mathcal{K}_i^l$ can be estimated by
using the chain rule, resulting in a back-propagation equation of the error signal from the top layer:
\begin{subequations}\label{backprop}
\begin{align}
 \mathcal{K}_i^l&=\delta_i^lf'(z_i^l),\\
 \delta_i^l&=\sum_{k}\mathcal{K}_k^{l+1}\frac{\partial z_k^{l+1}}{\partial h_i^l},
 \end{align}
\end{subequations}
where $\delta_i^l\equiv\frac{\partial\mathcal{C}}{\partial h_i^l}$, and $f'(\cdotp)$ denotes the derivative of the transfer function.

To proceed, we have to compute $\frac{\partial z_k^{l+1}}{\partial h_i^l}$ and $\frac{\partial z_i^{l+1}}{\partial\bm{\theta}_{ki}^l}$. The first derivative
characterizes how sensitive the pre-activation is under the change of the input activity of one neuron, and this response is computed as follows:
\begin{equation}\label{resp1}
 \frac{\partial z_k^{l+1}}{\partial h_i^l}=\frac{\mu_{ik}^l}{\sqrt{N_l}}+\frac{(\varrho_{ik}^l-(\mu_{ik}^l)^2)h_i^l}{N_l\Delta_k^{l+1}}\epsilon_k^{l+1}.
\end{equation}
The second derivative characterizes how sensitive the pre-activation is under the change of the hyper-parameters of the SaS distribution. Because
the mean and variance, $G_i^{l+1}$ and $\Delta_i^{l+1}$, is a function of the hyper-parameters, the second derivative for each hyper-parameter can be derived similarly as
follows:
\begin{subequations}\label{resp2}
\begin{align}
 \frac{\partial z_i^{l+1}}{\partial m_{ki}^{l}}&=\frac{(1-\pi_{ki}^{l})h_k^{l}}{\sqrt{N_l}}+\frac{\mu_{ki}^{l}\pi_{ki}^{l}}{N_l\Delta_{i}^{l+1}}(h_k^{l})^2\epsilon_i^{l+1},\\
 \frac{\partial z_i^{l+1}}{\partial \pi_{ki}^{l}}&=-\frac{m_{ki}^{l}h_k^{l}}{\sqrt{N_l}}-\frac{(2\pi_{ki}^{l}-1)(m_{ki}^{l})^2+\Xi_{ki}^{l}}{2N_l\Delta_{i}^{l+1}}(h_k^{l})^2\epsilon_i^{l+1},\label{derpi}\\
 \frac{\partial z_i^{l+1}}{\partial \Xi_{ki}^{l}}&=\frac{1-\pi_{ki}^{l}}{2N_l\Delta_{i}^{l+1}}(h_k^{l})^2\epsilon_i^{l+1}.
 \end{align}
\end{subequations}
Eq.~(\ref{resp1}) and Eq.~(\ref{resp2}) share the same form with the pre-activation $z_i^l$ [Eq.~(\ref{rep})], due to the reparametrization trick used to
handle the uncertainty of weights in the hypothesis space. Therefore, the learning process of our model naturally captures the fluctuation of the hypothesis space, highlighting the significant difference from
the standard BP which computes only a point estimate of the connection weights. In particular, if we enforce $\bm{\pi}=0$ and $\bm{\Xi}=0$, $\mathbf{m}$ becomes identical to the weight configuration,
thus our learning equations will immediately 
recover the standard BP algorithm~\cite{Back-1986}. Therefore, our learning protocol can be thought of as a generalized back-propagation (gBP) at the weight distribution level, or the candidate-network ensemble level.

Our framework is a cheap way to compute the posterior distribution of the weights given the data, which can be alternatively 
realized by a Bayesian inference where a variational free energy is commonly optimized through Monte-Carlo samplings~\cite{Blei-2017,Huang-2019data,Bald-2019}. The optimization of the variational free energy for 
deep neural networks is computationally challenging, especially for the SaS prior whose entropy has no analytic form as well.
In contrast, gBP stores two times more parameters than BP, while other computational steps are exactly the same;
the training time is affordable, as a typical deep network
training scales linearly with the number of parameters~\cite{Sejnowski-2020}. 

We remark that during learning, the spike mass $\pi$ should be clipped as ${\rm max}(0,{\rm min}(1,\pi))$, and the variance $\Xi\leftarrow{\rm max}(0,\Xi)$.
Learning the SaS model allows for credit assignment to each connection, considering a network ensemble realizing the same task. An effective network can be constructed by sampling the learned SaS distribution.

\begin{figure}
     \includegraphics[bb=5 4 1268 420,width=0.5\textwidth]{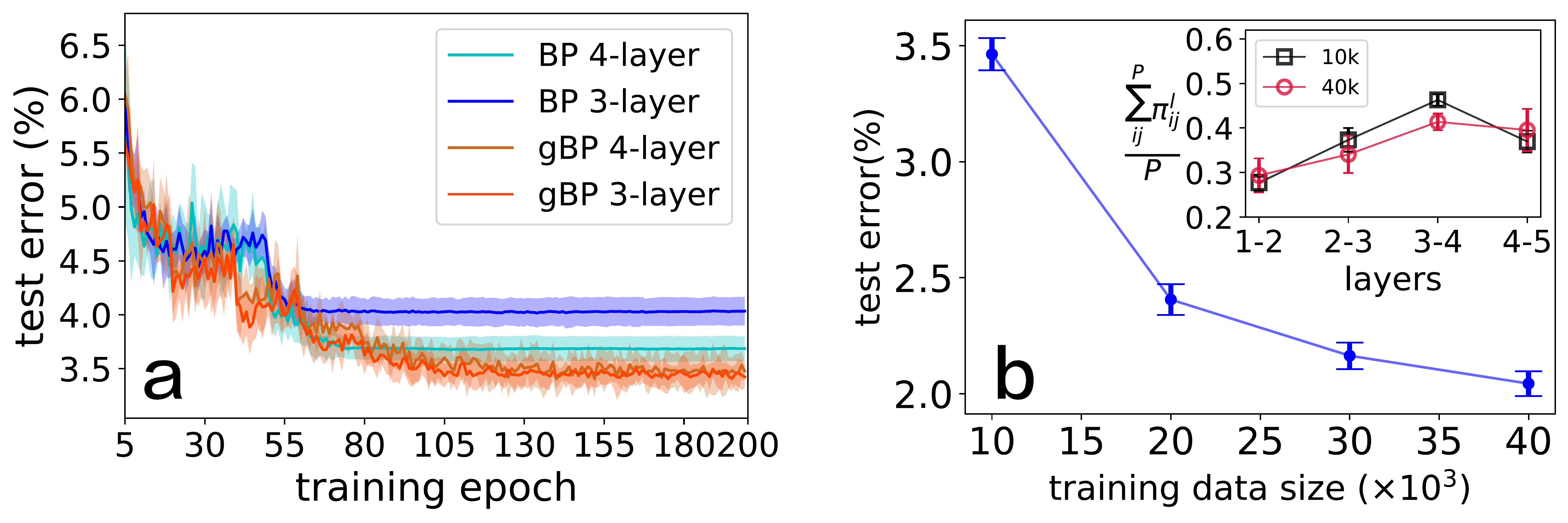}
  \caption{Properties of gBP in testing performances on unseen data.
  (a) Training trajectories of a network architecture of three or four layers. The architecture is defined as 784-100-100-10 (four layers) or 784-100-10 (three layers), where each number indicates
  the corresponding layer width. Networks are trained on the data size of $T=10^4$ images and tested on another unseen-data size of $V=10^4$ images. The fluctuation is computed from five independent runs.
  (b) Test errors of gBP versus training data size. The error bar characterizes the fluctuation across ten independently-sampled network architectures from the SaS distribution. The same
  network of four layers as in (a) is used. The inset shows the sparsity per connection as a function of layers, for which networks of five layers are used and the hidden layer width is still $100$
  nodes. Each marker is an average over ten independent runs. k in the inset means the training data size in the unit of $10^3$.
    }\label{Perf}
\end{figure}
The size of the hypothesis space can be approximated by the model entropy $S=-\int_{\mathbb{R}^{\mathcal{D}}}P(\bm{w})\ln P(\bm{w})d\bm{w}$, where $\mathcal{D}$ is
the number of weight parameters in the network.
Because the joint distribution of weights is assumed to be factorized across individual connections, $S=\sum_{\ell}S_{\ell}$, where the entropy
contribution of each individual connection ($\ell$) is expressed as~\cite{SM}
\begin{equation}\label{entropy}
 S_{\ell}=-\pi_{\ell}\ln\Biggl[\pi_{\ell}+(1-\pi_{\ell})\mathcal{N}(0|m_{\ell},\Xi_{\ell})\Biggr]-\frac{(1-\pi_{\ell})}{\mathcal{B}}\sum_s\Gamma(\epsilon_s),
\end{equation}
where $\mathcal{B}$ denotes the number of standard Gaussian random variables $\epsilon_s$, and $\Gamma(\epsilon_s)=\ln\Bigl[\pi_{\ell}\delta(m_{\ell}+\sqrt{\Xi_{\ell}}\epsilon_s)+\frac{(1-\pi_{\ell})}{\sqrt{\Xi_{\ell}}}\mathcal{N}(\epsilon_s|0,1)\Bigr]$.
If $\pi_{\ell}=0$, the entropy $S_{\ell}$ can be analytically computed as $\frac{1}{2}\ln(2\pi e\Xi_{\ell})$. If $\Xi_{\ell}=0$, the Gaussian distribution reduces to a Dirac delta function, and the entropy
becomes an entropy of discrete random variables.

\textit{Results.}---
Despite working at the synaptic weight distribution level, gBP can reach a similar even better
test accuracy than that of BP [Fig.~\ref{Perf} (a)]. As expected, the test error for gBP
decreases with training data size [Fig.~\ref{Perf} (b)]. The error bar implies that \textit{effective networks} sampled from the learned SaS distribution still yield accurate predictions, confirming the network ensemble assumption. 
Our theory
also reveals that the sparsity, obtained from the statistics of $\{\pi_{ij}^l\}$, grows first and then decreases [the inset of Fig.~\ref{Perf} (b)], suggesting that the actual working network
does not used up all synaptic connections, consistent with empirical observations of network compression~\cite{Han-2015,packnet-2017,Ticket-2019} and theoretical studies of toy models~\cite{Rap-2018,Saad-2019}.
Along hierarchical stages of the deep neural network, the initial stage is responsible for encoding, therefore the sparsity must be low to ensure that there is sufficient space for encoding the
important information, while at the middle stage, the encoded information is re-coded through hidden representations, suggesting that the noisy information is further distilled, which
may explain why the sparsity goes up. Finally, to guarantee the feature-selective information being extracted, the sparsity must drop to yield an accurate classification.
Therefore, our model can interpret the deep learning as an encoding-recoding-decoding process, as also argued in a biological neural hierarchy~\cite{Xaq-2017}.

\begin{figure}
     \includegraphics[bb=82 19 733 564,width=0.5\textwidth]{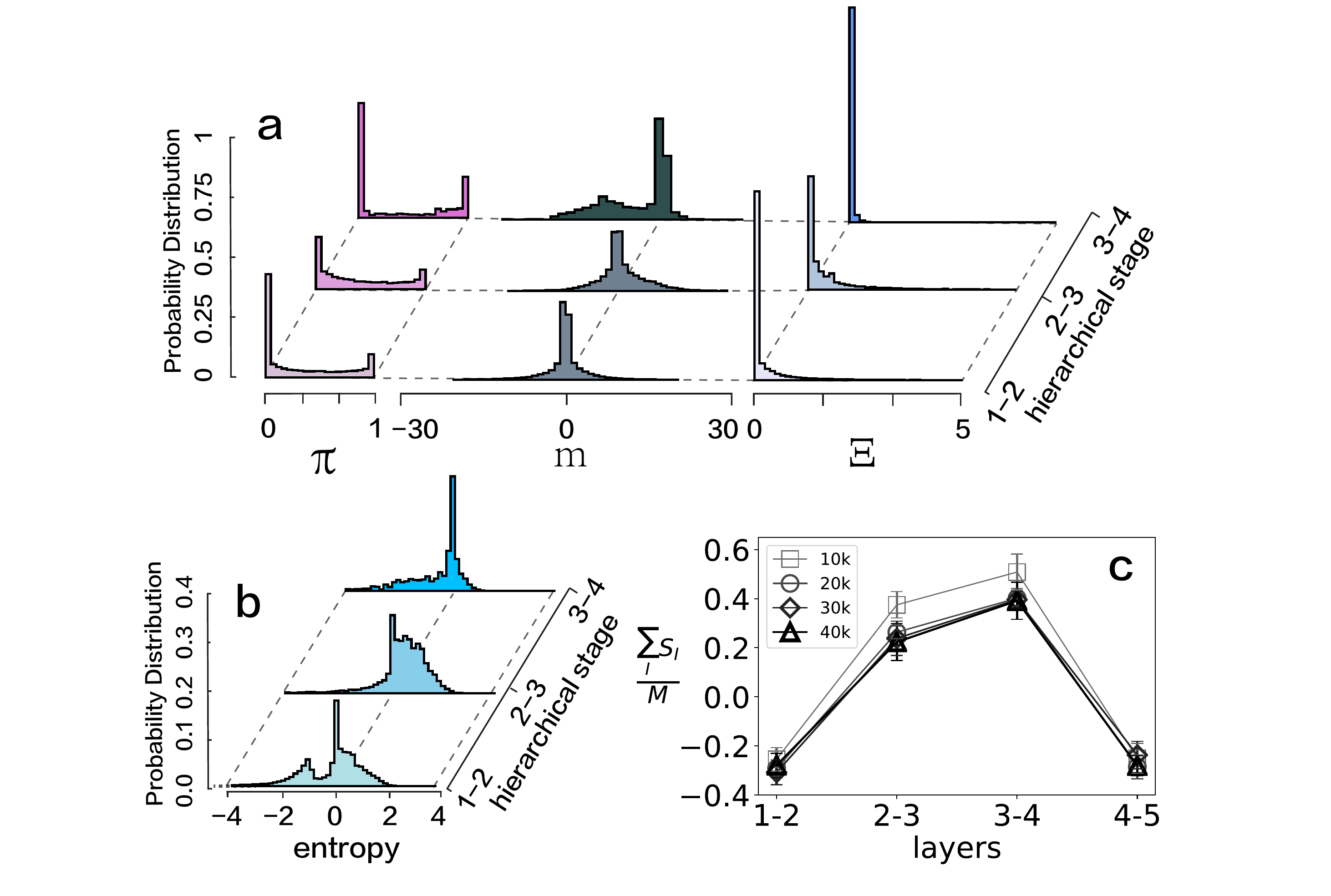}
  \caption{
  Statistical properties of trained four-layer networks with gBP. Unless stated otherwise, other training conditions are the same as in Fig.~\ref{Perf}.
  (a) Distribution of hyper-parameters $(\boldsymbol{\pi},\mathbf{m},\boldsymbol{\Xi})$ for a typical trained network.  
  (b) Distribution of connection entropy $S_{\ell}$ for a typical trained network.
  (c) Entropy per connection versus layers ($\mathcal{B}=100$). Different training data sizes are considered, and the result is averaged over ten independent runs of five-layer networks.
  }\label{Stat}
\end{figure}

To inspect more carefully what distribution of hyper-parameters gBP learned, we plot Fig.~\ref{Stat} (a) showing the evolution of the distribution across the hierarchical stages.
First, the spike mass $\bm{\pi}$ has a U-shaped distribution. One extreme is at $\pi=0$, suggesting that the corresponding connection carries feature-selective information and thus can not be
pruned, while the other extreme is at $\pi=1$, suggesting that the corresponding connection can be completely pruned. Apart from these two extremes, there exist a relatively small number of
connections which can be present or absent with certain probabilities. These connections may reflect nuisance factors in sensory inputs~\cite{Stef-2018}. 
The mean of the slab distribution has a relatively broad distribution, but the peak
is located around zero; an L-shaped distribution of the variance hyper-parameter is observed.
Note that if $\Xi=0$, the SaS distribution reduces to a Bernoulli distribution with two delta peaks. Due to Eq.~(\ref{derpi}), the point solution 
$(\boldsymbol{\pi}=0,\boldsymbol{\Xi}=0)$ is not a stable attractor for gBP~\cite{SM}. Moreover, as the training data size increases, the variance per connection displays 
a non-monotonic behavior with a residual value depending on the hierarchical stage~\cite{SM}. 

We can use the entropy $S_{\ell}$ for each connection to characterize the 
variability of the weight value. Note that the entropy can be negative for continuous random variables. The peak at zero [Fig.~\ref{Stat} (b)] indicates that 
a large number of connections are deterministic, including two cases: (i) $\pi=1$ (unimportant (UIP) weight); (ii) $\pi=0$ and $\Xi=0$ (very important (VIP) weight).
Fig.~\ref{Stat} (c) shows that the entropy per connection grows first, and then decreases. This non-monotonic behavior is the same with that of the sparsity in Fig.~\ref{Perf} (b).
The large entropy at the middle stage suggests that during the recoding process, the network has more degrees of freedom to manipulate the hypothesis space of the computational task~\cite{Huang-2020}.
Furthermore, more training data reduce the uncertainty (yet not to zero) until saturation [Fig.~\ref{Stat} (c)].

\begin{figure}
     \includegraphics[bb=4 5 1275 419,width=0.5\textwidth]{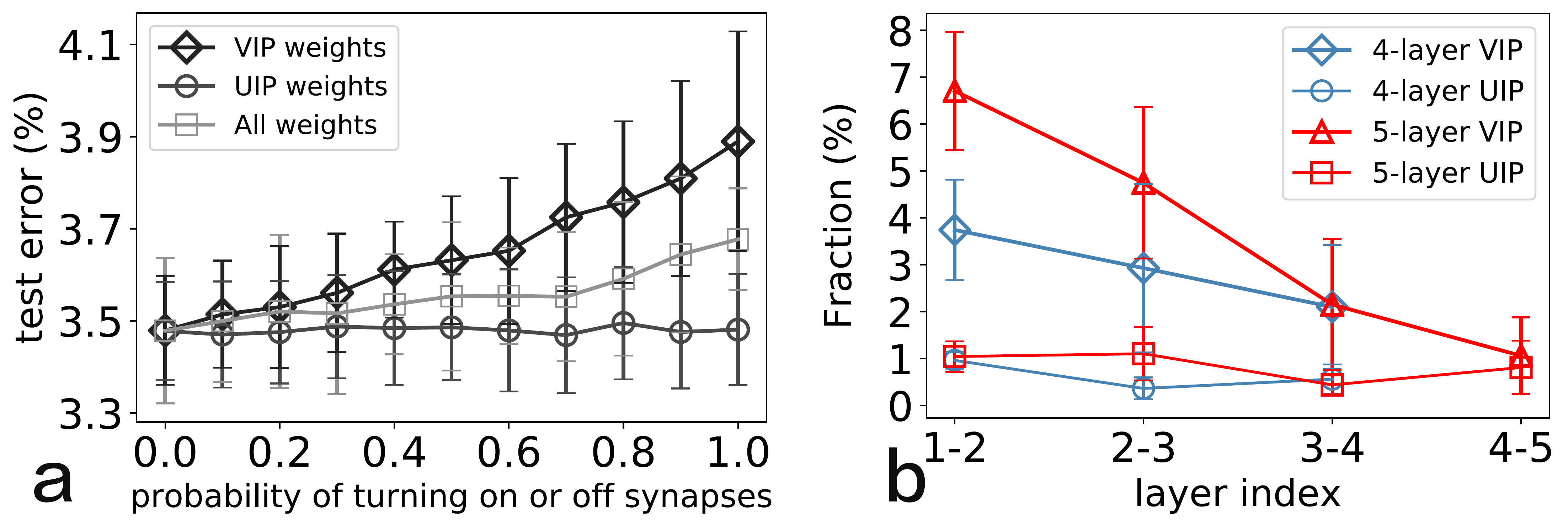}
  \caption{
 Effects of targeted-weight perturbation on test performances. Training conditions are the same as in Fig.~\ref{Perf}, and the result is averaged over ten independent runs.
 (a) VIP weights and all weights (without weight categorization) are stochastically turned off, keeping the same number, 
 while UIP weights are stochastically turned on and sampled from a standard Gaussian distribution. The perturbation is carried out at the first hierarchical stage of a four-layer network.
 (b) The fraction of VIP and UIP weights between neighboring layers.
  }\label{Prune}
\end{figure}

Finally, our mean-field framework can be used to explore effects of targeted-weight perturbation [Fig.~\ref{Prune} (a)], which previous studies of random deep models could not address~\cite{Rap-2018,Saad-2019}.
VIP weights play a significant role in determining the generalization capability, while turning on UIP weights does not impair the performance.
A random dilution of all weights behaves mildly in between. When going deeper, the perturbation effect becomes less evident (not shown), due to the less number of VIP weights [Fig.~\ref{Prune} (b)].

\textit{Conclusion.}---
In this work, we propose a statistical model of learning credit assignment in deep neural networks, resulting in an optimal random sub-network ensemble
explaining the behavior output of the hierarchical system.
The model can be solved by using mean-field methods, yielding a practical way to visualize consumed resources of connections and an ensemble of random weights---two key factors affecting the 
emergent decision making behavior of the network. Our framework can be applied to a more challenging CIFAR-$10$ dataset~\cite{Cifar-10} and obtains qualitatively similar results~\cite{SM}.
Our model thus provides deep insights towards understanding
many recent interesting empirical observations of random templates in both artificial and biological neural networks~\cite{LTH-2018,Hidden-2019,Zhou-2019,FSN-2019,Zador-2019}.
The model also provides a principled method for network compression saving memory and computation demands~\cite{Han-2015}, and could also have implications for continual learning of sequential tasks, where those important connections 
for old tasks are always protected to learn new tasks~\cite{Kirk-2017,packnet-2017}.  

As artificial neural networks become increasingly important to model the brain~\cite{DLNS-2019,FSN-2019},
our current study of artificial neural networks may provide cues for addressing how the brain solves the credit assignment problem.
Other promising directions include considering correlations among weights and generalization of this framework to 
the temporal credit assignment, where spatio-temporal information is considered in training recurrent neural networks~\cite{Werbos-1990,Suss-2009,Kording-2016}.


\begin{acknowledgments}
This research was supported by the start-up budget 74130-18831109 of the 100-talent-
program of Sun Yat-sen University, and the NSFC (Grant No. 11805284). 
\end{acknowledgments}


\begin{thebibliography}{40}
\expandafter\ifx\csname natexlab\endcsname\relax\def\natexlab#1{#1}\fi
\expandafter\ifx\csname bibnamefont\endcsname\relax
  \def\bibnamefont#1{#1}\fi
\expandafter\ifx\csname bibfnamefont\endcsname\relax
  \def\bibfnamefont#1{#1}\fi
\expandafter\ifx\csname citenamefont\endcsname\relax
  \def\citenamefont#1{#1}\fi
\expandafter\ifx\csname url\endcsname\relax
  \def\url#1{\texttt{#1}}\fi
\expandafter\ifx\csname urlprefix\endcsname\relax\def\urlprefix{URL }\fi
\providecommand{\bibinfo}[2]{#2}
\providecommand{\eprint}[2][]{\url{#2}}

\bibitem[{\citenamefont{Krizhevsky et~al.}(2012)\citenamefont{Krizhevsky,
  Sutskever, and Hinton}}]{Hinton-2012}
\bibinfo{author}{\bibfnamefont{A.}~\bibnamefont{Krizhevsky}},
  \bibinfo{author}{\bibfnamefont{I.}~\bibnamefont{Sutskever}},
  \bibnamefont{and} \bibinfo{author}{\bibfnamefont{G.~E.}
  \bibnamefont{Hinton}}, in \emph{\bibinfo{booktitle}{Advances in Neural
  Information Processing Systems 25}}, edited by
  \bibinfo{editor}{\bibfnamefont{F.}~\bibnamefont{Pereira}},
  \bibinfo{editor}{\bibfnamefont{C.~J.~C.} \bibnamefont{Burges}},
  \bibinfo{editor}{\bibfnamefont{L.}~\bibnamefont{Bottou}}, \bibnamefont{and}
  \bibinfo{editor}{\bibfnamefont{K.~Q.} \bibnamefont{Weinberger}}
  (\bibinfo{publisher}{Curran Associates, Inc.}, \bibinfo{year}{2012}), pp.
  \bibinfo{pages}{1097--1105}.

\bibitem[{\citenamefont{Hannun et~al.}(2014)\citenamefont{Hannun, Case, Casper,
  Catanzaro, Diamos, Elsen, Prenger, Satheesh, Sengupta, Coates
  et~al.}}]{Dsp-2014}
\bibinfo{author}{\bibfnamefont{A.}~\bibnamefont{Hannun}},
  \bibinfo{author}{\bibfnamefont{C.}~\bibnamefont{Case}},
  \bibinfo{author}{\bibfnamefont{J.}~\bibnamefont{Casper}},
  \bibinfo{author}{\bibfnamefont{B.}~\bibnamefont{Catanzaro}},
  \bibinfo{author}{\bibfnamefont{G.}~\bibnamefont{Diamos}},
  \bibinfo{author}{\bibfnamefont{E.}~\bibnamefont{Elsen}},
  \bibinfo{author}{\bibfnamefont{R.}~\bibnamefont{Prenger}},
  \bibinfo{author}{\bibfnamefont{S.}~\bibnamefont{Satheesh}},
  \bibinfo{author}{\bibfnamefont{S.}~\bibnamefont{Sengupta}},
  \bibinfo{author}{\bibfnamefont{A.}~\bibnamefont{Coates}},
  \bibnamefont{et~al.}, \bibinfo{journal}{arXiv:1412.5567}
  (\bibinfo{year}{2014}).

\bibitem[{\citenamefont{{He} et~al.}(2015)\citenamefont{{He}, {Zhang}, {Ren},
  and {Sun}}}]{He-2015}
\bibinfo{author}{\bibfnamefont{K.}~\bibnamefont{{He}}},
  \bibinfo{author}{\bibfnamefont{X.}~\bibnamefont{{Zhang}}},
  \bibinfo{author}{\bibfnamefont{S.}~\bibnamefont{{Ren}}}, \bibnamefont{and}
  \bibinfo{author}{\bibfnamefont{J.}~\bibnamefont{{Sun}}}, in
  \emph{\bibinfo{booktitle}{2015 IEEE International Conference on Computer
  Vision (ICCV)}} (\bibinfo{publisher}{IEEE Computer Society},
  \bibinfo{year}{2015}), pp. \bibinfo{pages}{1026--1034}.

\bibitem[{\citenamefont{Silver et~al.}(2017)\citenamefont{Silver,
  Schrittwieser, Simonyan, Antonoglou, Huang, Guez, Hubert, Baker, Lai, Bolton
  et~al.}}]{Silver-2017}
\bibinfo{author}{\bibfnamefont{D.}~\bibnamefont{Silver}},
  \bibinfo{author}{\bibfnamefont{J.}~\bibnamefont{Schrittwieser}},
  \bibinfo{author}{\bibfnamefont{K.}~\bibnamefont{Simonyan}},
  \bibinfo{author}{\bibfnamefont{I.}~\bibnamefont{Antonoglou}},
  \bibinfo{author}{\bibfnamefont{A.}~\bibnamefont{Huang}},
  \bibinfo{author}{\bibfnamefont{A.}~\bibnamefont{Guez}},
  \bibinfo{author}{\bibfnamefont{T.}~\bibnamefont{Hubert}},
  \bibinfo{author}{\bibfnamefont{L.}~\bibnamefont{Baker}},
  \bibinfo{author}{\bibfnamefont{M.}~\bibnamefont{Lai}},
  \bibinfo{author}{\bibfnamefont{A.}~\bibnamefont{Bolton}},
  \bibnamefont{et~al.}, \bibinfo{journal}{Nature}
  \textbf{\bibinfo{volume}{550}}, \bibinfo{pages}{354} (\bibinfo{year}{2017}).

\bibitem[{\citenamefont{Richards et~al.}(2019)\citenamefont{Richards,
  Lillicrap, Beaudoin, Bengio, Bogacz, Christensen, Clopath, Costa, de~Berker,
  Ganguli et~al.}}]{DLNS-2019}
\bibinfo{author}{\bibfnamefont{B.~A.} \bibnamefont{Richards}},
  \bibinfo{author}{\bibfnamefont{T.~P.} \bibnamefont{Lillicrap}},
  \bibinfo{author}{\bibfnamefont{P.}~\bibnamefont{Beaudoin}},
  \bibinfo{author}{\bibfnamefont{Y.}~\bibnamefont{Bengio}},
  \bibinfo{author}{\bibfnamefont{R.}~\bibnamefont{Bogacz}},
  \bibinfo{author}{\bibfnamefont{A.}~\bibnamefont{Christensen}},
  \bibinfo{author}{\bibfnamefont{C.}~\bibnamefont{Clopath}},
  \bibinfo{author}{\bibfnamefont{R.~P.} \bibnamefont{Costa}},
  \bibinfo{author}{\bibfnamefont{A.}~\bibnamefont{de~Berker}},
  \bibinfo{author}{\bibfnamefont{S.}~\bibnamefont{Ganguli}},
  \bibnamefont{et~al.}, \bibinfo{journal}{Nature Neuroscience}
  \textbf{\bibinfo{volume}{22}}, \bibinfo{pages}{1761} (\bibinfo{year}{2019}).

\bibitem[{\citenamefont{Sinz et~al.}(2019)\citenamefont{Sinz, Pitkow, Reimer,
  Bethge, and Tolias}}]{Neuron-2019}
\bibinfo{author}{\bibfnamefont{F.~H.} \bibnamefont{Sinz}},
  \bibinfo{author}{\bibfnamefont{X.}~\bibnamefont{Pitkow}},
  \bibinfo{author}{\bibfnamefont{J.}~\bibnamefont{Reimer}},
  \bibinfo{author}{\bibfnamefont{M.}~\bibnamefont{Bethge}}, \bibnamefont{and}
  \bibinfo{author}{\bibfnamefont{A.~S.} \bibnamefont{Tolias}},
  \bibinfo{journal}{Neuron} \textbf{\bibinfo{volume}{103}},
  \bibinfo{pages}{967} (\bibinfo{year}{2019}).

\bibitem[{\citenamefont{{Frankle} and {Carbin}}(2018)}]{LTH-2018}
\bibinfo{author}{\bibfnamefont{J.}~\bibnamefont{{Frankle}}} \bibnamefont{and}
  \bibinfo{author}{\bibfnamefont{M.}~\bibnamefont{{Carbin}}},
  \bibinfo{journal}{arXiv:1803.03635}  (\bibinfo{year}{2018}),
  \bibinfo{note}{in ICLR 2019}.

\bibitem[{\citenamefont{{Zhou} et~al.}(2019)\citenamefont{{Zhou}, {Lan}, {Liu},
  and {Yosinski}}}]{Zhou-2019}
\bibinfo{author}{\bibfnamefont{H.}~\bibnamefont{{Zhou}}},
  \bibinfo{author}{\bibfnamefont{J.}~\bibnamefont{{Lan}}},
  \bibinfo{author}{\bibfnamefont{R.}~\bibnamefont{{Liu}}}, \bibnamefont{and}
  \bibinfo{author}{\bibfnamefont{J.}~\bibnamefont{{Yosinski}}},
  \bibinfo{journal}{arXiv:1905.01067}  (\bibinfo{year}{2019}),
  \bibinfo{note}{in NeurIPS 2019}.

\bibitem[{\citenamefont{Ramanujan et~al.}(2019)\citenamefont{Ramanujan,
  Wortsman, Kembhavi, Farhadi, and Rastegari}}]{Hidden-2019}
\bibinfo{author}{\bibfnamefont{V.}~\bibnamefont{Ramanujan}},
  \bibinfo{author}{\bibfnamefont{M.}~\bibnamefont{Wortsman}},
  \bibinfo{author}{\bibfnamefont{A.}~\bibnamefont{Kembhavi}},
  \bibinfo{author}{\bibfnamefont{A.}~\bibnamefont{Farhadi}}, \bibnamefont{and}
  \bibinfo{author}{\bibfnamefont{M.}~\bibnamefont{Rastegari}},
  \bibinfo{journal}{arXiv:1911.13299}  (\bibinfo{year}{2019}).

\bibitem[{\citenamefont{Morcos et~al.}(2019)\citenamefont{Morcos, Yu, Paganini,
  and Tian}}]{Ticket-2019}
\bibinfo{author}{\bibfnamefont{A.~S.} \bibnamefont{Morcos}},
  \bibinfo{author}{\bibfnamefont{H.}~\bibnamefont{Yu}},
  \bibinfo{author}{\bibfnamefont{M.}~\bibnamefont{Paganini}}, \bibnamefont{and}
  \bibinfo{author}{\bibfnamefont{Y.}~\bibnamefont{Tian}},
  \bibinfo{journal}{arXiv:1906.02773}  (\bibinfo{year}{2019}),
  \bibinfo{note}{in NeurIPS 2019}.

\bibitem[{\citenamefont{Gaier and Ha}(2019)}]{Weight-2019}
\bibinfo{author}{\bibfnamefont{A.}~\bibnamefont{Gaier}} \bibnamefont{and}
  \bibinfo{author}{\bibfnamefont{D.}~\bibnamefont{Ha}},
  \bibinfo{journal}{arXiv:1906.04358}  (\bibinfo{year}{2019}),
  \bibinfo{note}{in NeurIPS 2019}.

\bibitem[{\citenamefont{Baek et~al.}(2019)\citenamefont{Baek, Song, Jang, Kim,
  and Paik}}]{FSN-2019}
\bibinfo{author}{\bibfnamefont{S.}~\bibnamefont{Baek}},
  \bibinfo{author}{\bibfnamefont{M.}~\bibnamefont{Song}},
  \bibinfo{author}{\bibfnamefont{J.}~\bibnamefont{Jang}},
  \bibinfo{author}{\bibfnamefont{G.}~\bibnamefont{Kim}}, \bibnamefont{and}
  \bibinfo{author}{\bibfnamefont{S.-B.} \bibnamefont{Paik}},
  \bibinfo{journal}{bioRxiv}  (\bibinfo{year}{2019}),
  \urlprefix\url{https://www.biorxiv.org/content/early/2019/11/29/857466}.

\bibitem[{mni()}]{mnist}
\bibinfo{note}{Y. LeCun, The MNIST database of handwritten digits, retrieved
  from http://yann.lecun.com/exdb/mnist.}

\bibitem[{\citenamefont{Nair and Hinton}(2010)}]{Nair-2010}
\bibinfo{author}{\bibfnamefont{V.}~\bibnamefont{Nair}} \bibnamefont{and}
  \bibinfo{author}{\bibfnamefont{G.~E.} \bibnamefont{Hinton}}, in
  \emph{\bibinfo{booktitle}{Proceedings of the 27th International Conference on
  International Conference on Machine Learning}}
  (\bibinfo{publisher}{Omnipress}, \bibinfo{address}{USA},
  \bibinfo{year}{2010}), ICML'10, pp. \bibinfo{pages}{807--814}.

\bibitem[{\citenamefont{Mitchell and Beauchamp}(1988)}]{sas-1988}
\bibinfo{author}{\bibfnamefont{T.~J.} \bibnamefont{Mitchell}} \bibnamefont{and}
  \bibinfo{author}{\bibfnamefont{J.~J.} \bibnamefont{Beauchamp}},
  \bibinfo{journal}{Journal of the American Statistical Association}
  \textbf{\bibinfo{volume}{83}}, \bibinfo{pages}{1023} (\bibinfo{year}{1988}).

\bibitem[{\citenamefont{Ishwaran and Rao}(2005)}]{Rao-2005}
\bibinfo{author}{\bibfnamefont{H.}~\bibnamefont{Ishwaran}} \bibnamefont{and}
  \bibinfo{author}{\bibfnamefont{J.~S.} \bibnamefont{Rao}},
  \bibinfo{journal}{Ann. Statist.} \textbf{\bibinfo{volume}{33}},
  \bibinfo{pages}{730} (\bibinfo{year}{2005}).

\bibitem[{\citenamefont{Han et~al.}(2015)\citenamefont{Han, Pool, Tran, and
  Dally}}]{Han-2015}
\bibinfo{author}{\bibfnamefont{S.}~\bibnamefont{Han}},
  \bibinfo{author}{\bibfnamefont{J.}~\bibnamefont{Pool}},
  \bibinfo{author}{\bibfnamefont{J.}~\bibnamefont{Tran}}, \bibnamefont{and}
  \bibinfo{author}{\bibfnamefont{W.~J.} \bibnamefont{Dally}},
  \bibinfo{journal}{arXiv:1506.02626}  (\bibinfo{year}{2015}),
  \bibinfo{note}{in NIPS 2015}.

\bibitem[{\citenamefont{{Ullrich} et~al.}(2017)\citenamefont{{Ullrich},
  {Meeds}, and {Welling}}}]{Max-2017}
\bibinfo{author}{\bibfnamefont{K.}~\bibnamefont{{Ullrich}}},
  \bibinfo{author}{\bibfnamefont{E.}~\bibnamefont{{Meeds}}}, \bibnamefont{and}
  \bibinfo{author}{\bibfnamefont{M.}~\bibnamefont{{Welling}}},
  \bibinfo{journal}{arXiv:1702.04008}  (\bibinfo{year}{2017}),
  \bibinfo{note}{in ICLR 2017}.

\bibitem[{\citenamefont{Parisi et~al.}(2019)\citenamefont{Parisi, Kemker, Part,
  Kanan, and Wermter}}]{CLL-2019}
\bibinfo{author}{\bibfnamefont{G.~I.} \bibnamefont{Parisi}},
  \bibinfo{author}{\bibfnamefont{R.}~\bibnamefont{Kemker}},
  \bibinfo{author}{\bibfnamefont{J.~L.} \bibnamefont{Part}},
  \bibinfo{author}{\bibfnamefont{C.}~\bibnamefont{Kanan}}, \bibnamefont{and}
  \bibinfo{author}{\bibfnamefont{S.}~\bibnamefont{Wermter}},
  \bibinfo{journal}{Neural Networks} \textbf{\bibinfo{volume}{113}},
  \bibinfo{pages}{54} (\bibinfo{year}{2019}).

\bibitem[{\citenamefont{Huang and Goudarzi}(2018)}]{Rap-2018}
\bibinfo{author}{\bibfnamefont{H.}~\bibnamefont{Huang}} \bibnamefont{and}
  \bibinfo{author}{\bibfnamefont{A.}~\bibnamefont{Goudarzi}},
  \bibinfo{journal}{Phys. Rev. E} \textbf{\bibinfo{volume}{98}},
  \bibinfo{pages}{042311} (\bibinfo{year}{2018}).

\bibitem[{\citenamefont{{McClure} and {Kriegeskorte}}(2016)}]{Krieg-2016}
\bibinfo{author}{\bibfnamefont{P.}~\bibnamefont{{McClure}}} \bibnamefont{and}
  \bibinfo{author}{\bibfnamefont{N.}~\bibnamefont{{Kriegeskorte}}},
  \bibinfo{journal}{arXiv:1611.01639}  (\bibinfo{year}{2016}).

\bibitem[{\citenamefont{Nowlan and Hinton}(1992)}]{Nowlan-1992}
\bibinfo{author}{\bibfnamefont{S.~J.} \bibnamefont{Nowlan}} \bibnamefont{and}
  \bibinfo{author}{\bibfnamefont{G.~E.} \bibnamefont{Hinton}},
  \bibinfo{journal}{Neural Computation} \textbf{\bibinfo{volume}{4}},
  \bibinfo{pages}{473} (\bibinfo{year}{1992}).

\bibitem[{\citenamefont{{Shayer} et~al.}(2017)\citenamefont{{Shayer}, {Levi},
  and {Fetaya}}}]{LRT-2017}
\bibinfo{author}{\bibfnamefont{O.}~\bibnamefont{{Shayer}}},
  \bibinfo{author}{\bibfnamefont{D.}~\bibnamefont{{Levi}}}, \bibnamefont{and}
  \bibinfo{author}{\bibfnamefont{E.}~\bibnamefont{{Fetaya}}},
  \bibinfo{journal}{arXiv:1710.07739}  (\bibinfo{year}{2017}),
  \bibinfo{note}{in ICLR 2018}.

\bibitem[{\citenamefont{Huang}(2020)}]{Huang-2019data}
\bibinfo{author}{\bibfnamefont{H.}~\bibnamefont{Huang}},
  \bibinfo{journal}{Phys. Rev. E} \textbf{\bibinfo{volume}{102}},
  \bibinfo{pages}{030301(R)} (\bibinfo{year}{2020}).

\bibitem[{SM()}]{SM}
\bibinfo{note}{See supplemental material at phys rev lett for derivations of the
  entropy formula, transformation, and more simulation results related to the
  MNIST amd CIFAR-10 datasets.}

\bibitem[{\citenamefont{{Rumelhart David E.}
  et~al.}(1986)\citenamefont{{Rumelhart David E.}, {Hinton Geoffrey E.}, and
  {Williams Ronald J.}}}]{Back-1986}
\bibinfo{author}{\bibnamefont{{Rumelhart David E.}}},
  \bibinfo{author}{\bibnamefont{{Hinton Geoffrey E.}}}, \bibnamefont{and}
  \bibinfo{author}{\bibnamefont{{Williams Ronald J.}}},
  \bibinfo{journal}{Nature} \textbf{\bibinfo{volume}{323}},
  \bibinfo{pages}{533} (\bibinfo{year}{1986}).

\bibitem[{\citenamefont{Blei et~al.}(2017)\citenamefont{Blei, Kucukelbir, and
  McAuliffe}}]{Blei-2017}
\bibinfo{author}{\bibfnamefont{D.~M.} \bibnamefont{Blei}},
  \bibinfo{author}{\bibfnamefont{A.}~\bibnamefont{Kucukelbir}},
  \bibnamefont{and} \bibinfo{author}{\bibfnamefont{J.~D.}
  \bibnamefont{McAuliffe}}, \bibinfo{journal}{Journal of the American
  Statistical Association} \textbf{\bibinfo{volume}{112}}, \bibinfo{pages}{859}
  (\bibinfo{year}{2017}).

\bibitem[{\citenamefont{{Baldock} and {Marzari}}(2019)}]{Bald-2019}
\bibinfo{author}{\bibfnamefont{R.~J.~N.} \bibnamefont{{Baldock}}}
  \bibnamefont{and}
  \bibinfo{author}{\bibfnamefont{N.}~\bibnamefont{{Marzari}}},
  \bibinfo{journal}{arXiv:1904.04154}  (\bibinfo{year}{2019}).

\bibitem[{\citenamefont{Sejnowski}(2020)}]{Sejnowski-2020}
\bibinfo{author}{\bibfnamefont{T.~J.} \bibnamefont{Sejnowski}},
  \bibinfo{journal}{Proceedings of the National Academy of Sciences}
  (\bibinfo{year}{2020}).

\bibitem[{\citenamefont{Mallya and Lazebnik}(2018)}]{packnet-2017}
\bibinfo{author}{\bibfnamefont{A.}~\bibnamefont{Mallya}} \bibnamefont{and}
  \bibinfo{author}{\bibfnamefont{S.}~\bibnamefont{Lazebnik}}, in
  \emph{\bibinfo{booktitle}{The IEEE Conference on Computer Vision and Pattern
  Recognition (CVPR)}} (\bibinfo{publisher}{IEEE Computer Society},
  \bibinfo{year}{2018}), pp. \bibinfo{pages}{7765--7773}.

\bibitem[{\citenamefont{Li and Saad}(2020)}]{Saad-2019}
\bibinfo{author}{\bibfnamefont{B.}~\bibnamefont{Li}} \bibnamefont{and}
  \bibinfo{author}{\bibfnamefont{D.}~\bibnamefont{Saad}},
  \bibinfo{journal}{Journal of Physics A: Mathematical and Theoretical}
  \textbf{\bibinfo{volume}{53}}, \bibinfo{pages}{104002}
  (\bibinfo{year}{2020}).

\bibitem[{\citenamefont{Pitkow and Angelaki}(2017)}]{Xaq-2017}
\bibinfo{author}{\bibfnamefont{X.}~\bibnamefont{Pitkow}} \bibnamefont{and}
  \bibinfo{author}{\bibfnamefont{D.~E.} \bibnamefont{Angelaki}},
  \bibinfo{journal}{Neuron} \textbf{\bibinfo{volume}{94}}, \bibinfo{pages}{943}
  (\bibinfo{year}{2017}).

\bibitem[{\citenamefont{Achille and Soatto}(2018)}]{Stef-2018}
\bibinfo{author}{\bibfnamefont{A.}~\bibnamefont{Achille}} \bibnamefont{and}
  \bibinfo{author}{\bibfnamefont{S.}~\bibnamefont{Soatto}},
  \bibinfo{journal}{Journal of Machine Learning Research}
  \textbf{\bibinfo{volume}{19}}, \bibinfo{pages}{1} (\bibinfo{year}{2018}).

\bibitem[{\citenamefont{Zou and Huang}(2020)}]{Huang-2020}
\bibinfo{author}{\bibfnamefont{W.}~\bibnamefont{Zou}} \bibnamefont{and}
  \bibinfo{author}{\bibfnamefont{H.}~\bibnamefont{Huang}},
  \bibinfo{journal}{arXiv:2007.08093}  (\bibinfo{year}{2020}).

\bibitem[{\citenamefont{Krizhevsky}(2009)}]{Cifar-10}
\bibinfo{author}{\bibfnamefont{A.}~\bibnamefont{Krizhevsky}},
  \bibinfo{type}{Tech. Rep.}, \bibinfo{institution}{University of Toronto,
  Toronto} (\bibinfo{year}{2009}), \bibinfo{note}{learning multiple layers of
  features from tiny images}.

\bibitem[{\citenamefont{Zador}(2019)}]{Zador-2019}
\bibinfo{author}{\bibfnamefont{A.~M.} \bibnamefont{Zador}},
  \bibinfo{journal}{Nature Communications} \textbf{\bibinfo{volume}{10}},
  \bibinfo{pages}{3770} (\bibinfo{year}{2019}).

\bibitem[{\citenamefont{Kirkpatrick et~al.}(2017)\citenamefont{Kirkpatrick,
  Pascanu, Rabinowitz, Veness, Desjardins, Rusu, Milan, Quan, Ramalho,
  Grabska-Barwinska et~al.}}]{Kirk-2017}
\bibinfo{author}{\bibfnamefont{J.}~\bibnamefont{Kirkpatrick}},
  \bibinfo{author}{\bibfnamefont{R.}~\bibnamefont{Pascanu}},
  \bibinfo{author}{\bibfnamefont{N.}~\bibnamefont{Rabinowitz}},
  \bibinfo{author}{\bibfnamefont{J.}~\bibnamefont{Veness}},
  \bibinfo{author}{\bibfnamefont{G.}~\bibnamefont{Desjardins}},
  \bibinfo{author}{\bibfnamefont{A.~A.} \bibnamefont{Rusu}},
  \bibinfo{author}{\bibfnamefont{K.}~\bibnamefont{Milan}},
  \bibinfo{author}{\bibfnamefont{J.}~\bibnamefont{Quan}},
  \bibinfo{author}{\bibfnamefont{T.}~\bibnamefont{Ramalho}},
  \bibinfo{author}{\bibfnamefont{A.}~\bibnamefont{Grabska-Barwinska}},
  \bibnamefont{et~al.}, \bibinfo{journal}{Proceedings of the National Academy
  of Sciences} \textbf{\bibinfo{volume}{114}}, \bibinfo{pages}{3521}
  (\bibinfo{year}{2017}).

\bibitem[{\citenamefont{{Werbos}}(1990)}]{Werbos-1990}
\bibinfo{author}{\bibfnamefont{P.~J.} \bibnamefont{{Werbos}}},
  \bibinfo{journal}{Proceedings of the IEEE} \textbf{\bibinfo{volume}{78}},
  \bibinfo{pages}{1550} (\bibinfo{year}{1990}).

\bibitem[{\citenamefont{Sussillo and Abbott}(2009)}]{Suss-2009}
\bibinfo{author}{\bibfnamefont{D.}~\bibnamefont{Sussillo}} \bibnamefont{and}
  \bibinfo{author}{\bibfnamefont{L.}~\bibnamefont{Abbott}},
  \bibinfo{journal}{Neuron} \textbf{\bibinfo{volume}{63}}, \bibinfo{pages}{544}
  (\bibinfo{year}{2009}).

\bibitem[{\citenamefont{Marblestone et~al.}(2016)\citenamefont{Marblestone,
  Wayne, and Kording}}]{Kording-2016}
\bibinfo{author}{\bibfnamefont{A.~H.} \bibnamefont{Marblestone}},
  \bibinfo{author}{\bibfnamefont{G.}~\bibnamefont{Wayne}}, \bibnamefont{and}
  \bibinfo{author}{\bibfnamefont{K.~P.} \bibnamefont{Kording}},
  \bibinfo{journal}{Front. Comput. Neurosci.} \textbf{\bibinfo{volume}{10}},
  \bibinfo{pages}{94} (\bibinfo{year}{2016}).

\end{thebibliography}


\end{document}